\documentclass[10pt,twocolumn,letterpaper]{article}

\usepackage[pagenumbers]{cvpr} 

\usepackage{amsmath,amssymb,amsfonts}
\usepackage{graphicx}
\usepackage{booktabs}
\usepackage{multirow}
\usepackage{xcolor}
\usepackage{subcaption}
\usepackage{algorithm}
\usepackage{algorithmic}
\usepackage{xspace}
\usepackage{microtype}
\usepackage{enumitem}
\usepackage{calc}
\usepackage{threeparttable}
\newcommand{\ours}{CAST\xspace}
\newcommand{\casa}{CASA\xspace}

\definecolor{cvprblue}{rgb}{0.21,0.49,0.74}
\usepackage[pagebackref,breaklinks,colorlinks,allcolors=cvprblue]{hyperref}

\def\paperID{29}
\def\confName{CVPR}
\def\confYear{2026}

\title{CAST: Channel-Aware Spatial Transfer Learning with Pseudo-Image Radar for Sign Language Recognition}

\author{
Md. Shakhoyat Rahman Shujon\textsuperscript{1},
Sheikh Md. Galib Mahim\textsuperscript{1},
Md. Milon Islam\textsuperscript{1},
Md Rezwanul Haque\textsuperscript{2},\\
Md Rabiul Islam\textsuperscript{3},
Hamdi Altaheri\textsuperscript{4},
Fakhri Karray\textsuperscript{2,5}
\\
\textsuperscript{1}Department of Computer Science and Engineering, Khulna University of Engineering \& Technology\\
\textsuperscript{2}Department of Electrical and Computer Engineering, University of Waterloo\\
\textsuperscript{3}Department of Electrical and Computer Engineering, Texas A\&M University\\
\textsuperscript{4}College of Applied Computer Science, King Saud University\\
\textsuperscript{5}Department of Machine Learning, Mohamed bin Zayed University of
Artificial Intelligence \\
{\tt\small\textsuperscript{1}skt104.shujon@gmail.com, \textsuperscript{1}galibmahim01@gmail.com, \textsuperscript{1}milonislam@cse.kuet.ac.bd,} \\
{\tt\small\textsuperscript{2}rezwan@uwaterloo.ca, 
\textsuperscript{3}rabiul\_islam@tamu.edu, \textsuperscript{4}haltaheri@ksu.edu.sa, \textsuperscript{2,5}karray@uwaterloo.ca} 
}

\begin{document}
\maketitle


\begingroup
\renewcommand\thefootnote{}
\vspace{-1em} 


\footnotetext{
\textsuperscript{©}\textit{Proceedings of the IEEE/CVF Conference on Computer Vision and Pattern Recognition (CVPR), MSLR Workshop @ CVPR 2026 in Denver (Colorado, USA). Copyright 2026 by the author(s).}
}

\endgroup

 \begin{abstract}
We propose CAST, a dual-stream architecture that utilizes channel-aware spatial transfer learning for isolated sign language recognition addressing the challenges of magnitude-only 60~GHz radar Range-Time Maps (RTM). The proposed framework combines three physics-aware architectures with pretrained vision backbones, which operate under radar-only constraints across clinical and alphabetical gestures. First, an explicit decibel-to-linear inversion is combined with a windowed fast Fourier transform that extracts Cadence Velocity Diagrams (CVD) while avoiding the harmonic artifacts that arise from the spectral analysis of log-compressed signals. Second, a cross-antenna spatial attention module applies attention to raw antenna channels before the convolution, preserving inter-receiver amplitude covariance. Third, an asymmetric cross-attention mechanism fuses representations from parallel ConvNeXt-Tiny (CVD) and EfficientNetV2-S (RTM) backbones. Extensive experiments reveal that the architecture achieves a Top-1 accuracy of 80.5\% under 5-fold cross-validation, establishing a 3.3\% improvement over the best single-model baseline (77.2\%). The findings suggest that physics-aware signal representations form a promising direction for radar-only sign language recognition under constrained sensor modalities. The source code is available at: \url{https://github.com/Shakhoyat/CAST-at-SignEval2026}.
\end{abstract}

\section{Introduction}
\label{sec:intro}

Vision-based sign language recognition has achieved relatively good performance in
controlled laboratory environments, with recent multimodal systems reporting
over 99\% accuracy on the Italian sign language vocabulary of the
MultiMeDaLIS benchmark~\cite{juranek2025multimodal}. In vision-based systems, the most challenging issue is that performance degrades significantly when the camera is removed. Clinical environments illustrate this problem clearly. Hospitals routinely handle patients who are deaf or
hard-of-hearing, but using cameras for patients raises regulation and consent issues that are difficult to resolve~\cite{mineo2024sign}.
Millimeter-wave radar provides a privacy-preserving solution. It captures the kinematics of hand and arm motion while inherently anonymizing the signer by operating
at 60\, GHz. Hence, it is appropriate for privacy-sensitive applications where continuous visual capture is difficult.

The CVPR 2026 MSLR workshop challenge (Track 2)
turns this scenario into a benchmark ~\cite{mineo2026benchmark,hasanaath2026signeval}. The
MultiMeDaLIS dataset ~\cite{mineo2024sign,caligiore2024multisource} provides 126 Italian sign
language classes (100 medical terms and 26 alphabet letters)
captured with an Infineon BGT60TR13C 60 GHz FMCW
radar across 205 sessions. In this setup, only Range-Time Maps (RTMs)
are available for evaluation; no video, depth, or complex-valued
radar data are used.

To address these constraints, the proposed approach in this challenge converts RTMs into
three-channel arrays and applies pretrained 2-D Convolutional Neural Networks (CNNs). The key point is that it removes two physical properties of the RTM that are important for recognition. The physical properties are as follows:

\begin{enumerate}[leftmargin=*,nosep]
\item \textbf{Velocity blindness:} The temporal axis in RTM encodes
    chronological kinematics to represent motion over time, not space. When this is treated as a static image, the temporal dimension is reduced, and important information, including velocity and periodicity, disappears, which helps to distinguish similar signs.
\item \textbf{Antenna-geometry ignorance:} The radar module places three
    receive antennas in an L-shape (two azimuth, one elevation). Using
    them as RGB channels (RX1, RX2, and RX3) ignores the geometric information and spacing between antennas.
\end{enumerate}

\paragraph{Contributions:}
The contribution of this work is to focus on physics-aware
architectures of existing methods, designed for magnitude-only radar data. The proposed architecture, Channel-Aware Spatial Transfer (CAST) learning, addresses the above gaps
through three modules:
\begin{itemize}[leftmargin=*,nosep]
\item \textbf{CVD extraction with dB-to-linear inversion:} We generate a
    Cadence Velocity Diagram (CVD) from magnitude-only RTMs by inverting
    the dB to a linear scale. Then, we apply a Blackman-Harris window Fast Fourier Transform (FFT)
    along the temporal axis. The linearization is necessary because
    applying a Fourier transform to log-scale data generates harmonic
     artifacts with no physical interpretation. This is due to the logarithm
    transformation violating the linear superposition assumption of Fourier
    analysis.
\item \textbf{Cross-Antenna Spatial Attention (CASA):} CASA encodes each
    antenna independently rather than treating
    three antenna channels as RGB-like inputs, stacks them into a sequence, and
    applies multi-head self-attention across antenna
    positions. The module processes raw antenna
    signals before the first convolution, preserving
    inter-receiver amplitude covariance.
\item \textbf{Asymmetric cross-attention fusion:} The RTM stream is utilized as
    the query, and the CVD stream provides keys and values. This allows the model to selectively retrieve velocity information when the structural range is insufficient for recognition.
\end
{itemize}
A controlled evaluation protocol is used to isolate the contribution of
each architectural module. Under the same single-model setting, CAST outperforms the baseline by 3.3\%
  (80.5\% vs.\ 77.2\%). The difference between
single-model and overall evaluation is discussed later. Furthermore, we analyze the practical limitations
caused by the 13\,fps capture rate and failure modes due to
sensor physics.

\section{Related Work}
\label{sec:related}

\paragraph{Radar-based gesture and sign language recognition:}
Fine-grained gesture sensing with 60\, GHz mmWave Frequency Modulated Continuous Wave (FMCW) radar was
established by Project Soli~\cite{lien2016soli}, which demonstrated that micro-scale hand movements generate
distinctive micro-Doppler signatures at millimetre wavelengths.
Early radar gesture frameworks converted micro-Doppler spectrograms or
Range-Doppler Maps (RDMs) to image tensors and applied 2-D CNNs pretrained on
natural images~\cite{surveyfmcw2023} . However, this introduces a 
representational mismatch as ImageNet kernels capture
visual textures rather than phase-based frequency modulations in radar signals. More recent radar-based architectures address this by employing hybrid
CNN-LSTM networks that separate spatial convolution from temporal sequence
modeling~\cite{wang2020novel} , and with Multi-view
De-interference Transformers that separate
gestures from background noise~\cite{rodar2024} .

For sign language, the TRACE
architecture~\cite{mineo2025trace} is the most closely
related prior work. TRACE employs a residual autoencoder to reduce
$128{\times}1024$ RDMs to a 256-D bottleneck, followed by a
six-layer, eight-head Transformer classifier, achieving 93.6\% accuracy on the same
126-class vocabulary. A related text-aligned
variant~\cite{mineo2025textaligned} further exploits
language supervision for radar representations. However, both works use full complex-valued Range-Doppler data at
significantly higher resolutions and frame rates than the RTMs available in our
challenge. Therefore, a direct accuracy comparison is not meaningful.
The authors of \cite{juranek2025multimodal,islam2025fusionensemble}
demonstrated that late fusion of radar
and video logits can degrade accuracy when the modalities have large gaps, confirming that radar-only recognition requires a customized
architecture rather than generic fusion methods.

To our knowledge, no directly comparable RTM-only baseline for 60\,GHz radar
sign language recognition has been published outside the MSLR workshop series. Existing magnitude-only radar gesture studies are limited to vocabularies of
16 or fewer classes, and all systems reporting above 96\% accuracy rely on
multi-modal fusion with Doppler or angle-of-arrival information.
Prior radar-only sign language work assumes access to complex-valued
Range-Doppler data at higher frame rates.
\vspace{-1em}
\paragraph{Pseudo-Doppler extraction from magnitude data:}
Standard Doppler extraction requires complex-valued radar data.
When only magnitude is available, temporal spectral analysis of the
range-profile time series can still recover the cadence of periodic
motions~\cite{kim2009microdoppler}. In literature, FFT-based analysis of magnitude signals is already used
in gait and activity recognition to extract micro-Doppler-like patterns from amplitude
observations~\cite{kim2009microdoppler}. However, our contribution is more specific: we demonstrate that applying
the FFT directly to dB-compressed data introduces harmonic artifacts
with no physical interpretation, and that performing a dB-to-linear inversion prior to FFT is a minimal correction required to preserve physically interpretable
cadence information. 

To our knowledge, this first-order approximation step has not
been validated in radar sign language recognition, where the input is
restricted to magnitude-based RTMs rather than full RDMs.
The kinematic envelope of RDMs carries dynamic signatures
that complement static spatial features~\cite{Sazonov_2025_ICCV}. In the proposed architecture, this information is not available; hence, accurate spectral extraction from magnitude data becomes the key design issue compared to post-hoc feature selection.

\begin{figure*}[t]
\centering
\includegraphics[width=\textwidth]{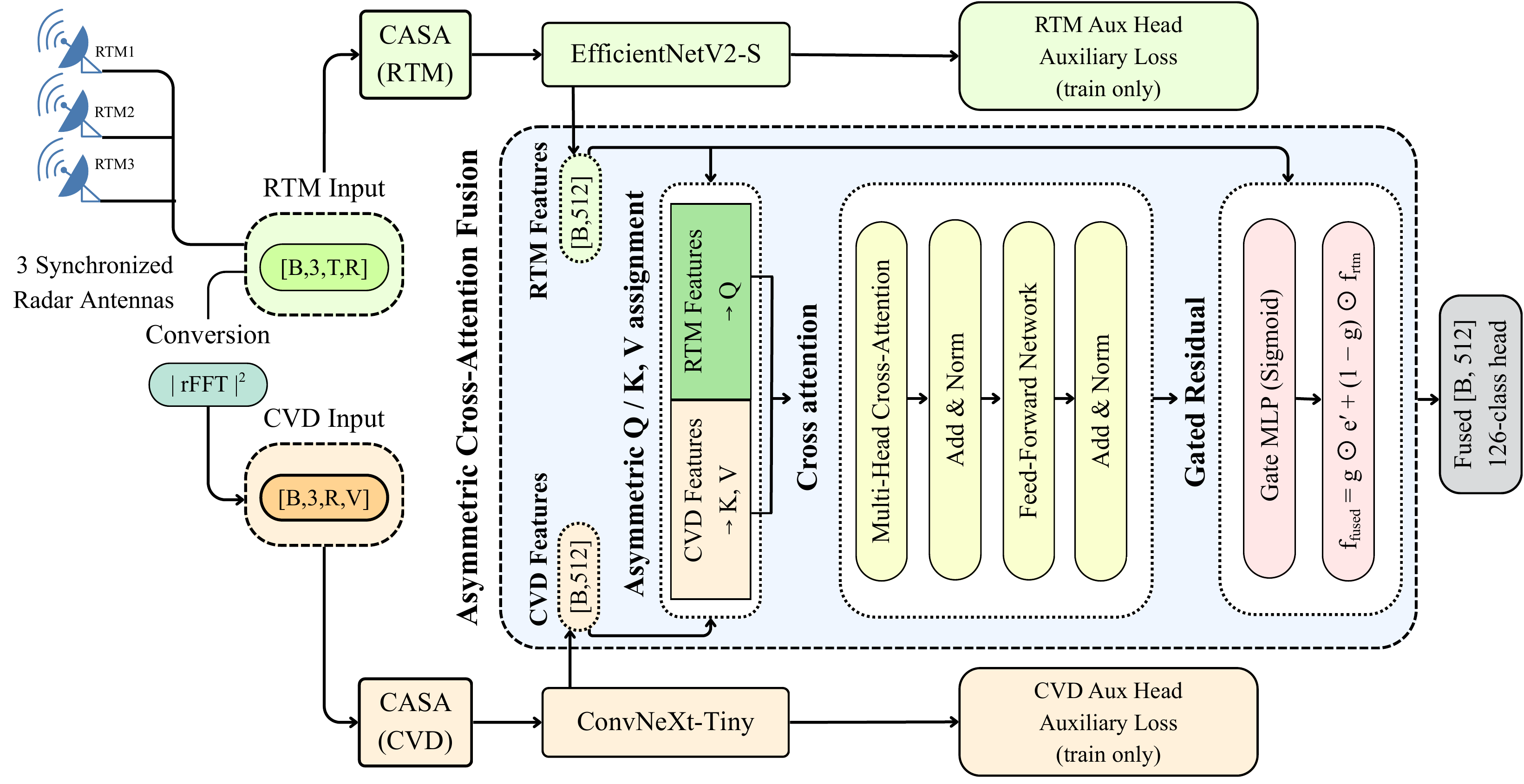}
\caption{%
Overall architecture of the proposed CAST architecture. Three-receiver RTMs are processed through per-stream CASA modules and fed
into EfficientNetV2-S (RTM) and ConvNeXt-Tiny (CVD) backbones.
An asymmetric cross-attention fusion module fuses the dual
representations, with a gated residual ensuring fallback to RTM-only
features when the CVD stream provides no meaningful information.
}
\label{fig:architecture}
\end{figure*}

\section{Method}
\label{sec:method}
The overall architecture of our proposed system is illustrated in 
Fig.~\ref{fig:architecture}. There are two parallel branches in this framework:
a top stream designed for normalized magnitude RTMs and a
bottom stream that extracts CVD.
Both streams pass the multi-receiver radar sequences through a CASA module and are then passed through their respective EfficientNetV2-S and ConvNeXt-Tiny backbones for feature extraction. RTM and CVD data are filtered and combined using a cross-attention fusion module, in which the RTM stream selectively retrieves additional information from the CVD streams. This fused information is then used for final recognition.

\subsection{Cadence Velocity Diagram Extraction}
\label{sec:cvd}

The MultiMeDaLIS RTMs are distributed as float32 arrays,
$\mathbf{R}_\mathrm{dB}\in\mathbb{R}^{T\times256}$ representing
$20\log_{10}(\text{amplitude})$, where $T$ is the number of slow-time
frames ($\approx$\,20--40 at 13\,fps) and 256 is the number of positive
range bins (indexed by $r$).  
\vspace{-1em}
\paragraph{Necessity of decibel-to-linear inversion:} A Fourier transform applied directly to logarithmic dB data is mathematically inconsistent. The logarithm converts a multiplicative
modulation of the carrier by a sinusoidal Doppler envelope into an additive form, which
violates the linearity assumption required for valid Fourier analysis.
An ideal sinusoidal modulation of amplitude $A(1 + m\cos(2\pi f_0 t))$
becomes $\log A + \log(1 + m\cos(2\pi f_0 t))$ after log compression.
This transformation does not generate a single spectral peak at $f_0$, it generates an
infinite series of harmonics, distorting the frequency axis. Expanding $\log(1 + m\cos(2\pi f_0 t))$ as a Taylor series for $|m| < 1$ generates
$m\cos(2\pi f_0 t) - \tfrac{m^2}{2}\cos^2(2\pi f_0 t) + \cdots$,
which contains terms at $2f_0$, $3f_0$, and all higher harmonics.
These higher-order terms generate artificial frequencies that do not correspond to the signer's actual physical kinematics. This effect is not only theoretical; in practice,
applying the FFT directly to dB-scale RTMs creates false peaks at
integer multiples of the true cadence frequency, leading the model to learn physically invalid features. Prior
magnitude-based cadence analysis in other domains mostly operates on
linear amplitude or on datasets where the log compression is limited. The full dB dynamic range available in the MultiMeDaLIS RTMs ($>$40\,dB)
amplifies the artifacts, resulting in a 1.7\% drop in accuracy (Table~\ref{tab:ablation}).
The first-order approximation recovers the original amplitude modulation, as shown in (1).
\begin{equation}
    \mathbf{R}_\mathrm{lin} = 10^{\mathbf{R}_\mathrm{dB}/20}
    \label{eq:linearise}
\end{equation}
\paragraph{Windowed FFT:}
A Blackman-Harris window $w[n]$ is applied along the temporal axis.
Blackman-Harris achieves $>$92\,dB sidelobe suppression, which is essential because dominant torso reflections (broad peaks in range) can mask weaker cadence signals created by hand motion. The window length $T$ (zero-padded to $N_\mathrm{FFT}=128$),
and the positive-frequency output is calculated in (2).
\begin{equation}
\begin{split}
    \mathbf{C}[k,r] = \Bigg|\,
      \sum_{n=0}^{N_{\mathrm{FFT}}-1}
      &\mathbf{R}_{\mathrm{lin}}[n,r]\, w[n] \\
      &\times \exp\!\left(-j\frac{2\pi k n}{N_{\mathrm{FFT}}}\right)
    \,\Bigg|
\end{split}
    \label{eq:fft}
\end{equation}
where, $n$ is the discrete-time index and
$k = 1,\,\ldots,\,\tfrac{N_{\mathrm{FFT}}}{2}$ indexes the positive
frequency bins.
\vspace{-1em}
\paragraph{CVD formation:}
The $k=0$ bin is discarded and
the magnitude is converted back to dB for dynamic-range compression as shown in (3).
\begin{equation}
    \mathrm{CVD}[r,k] = 20\log_{10}\!\left(\mathbf{C}[k,r]+\epsilon\right),
    \quad \epsilon=10^{-10}
    \label{eq:cvd}
\end{equation}
The resulting CVD has shape $256\times64$ per antenna. Given a frame rate of 13 fps, the
Nyquist limit is $\approx$6.5\,Hz, which is sufficient to capture the typical repetition rates of sign language gestures ($\approx$1--4\,Hz). Zero-padding interpolates the spectral bins
without increasing actual resolution, resulting in smoother feature maps for
convolutional processing. In preliminary experiments, a Continuous Wavelet Transform (CWT) scalogram achieved comparable performance but did not outperform the
FFT-based CVD (80.2\% vs.\ 80.5\% ), while incurring significantly higher computational cost; hence, we retain the FFT-based approach. 

\subsection{Cross-Antenna Spatial Attention}
\label{sec:casa}

\begin{figure}[t]
  \centering
  \includegraphics[width=\linewidth]{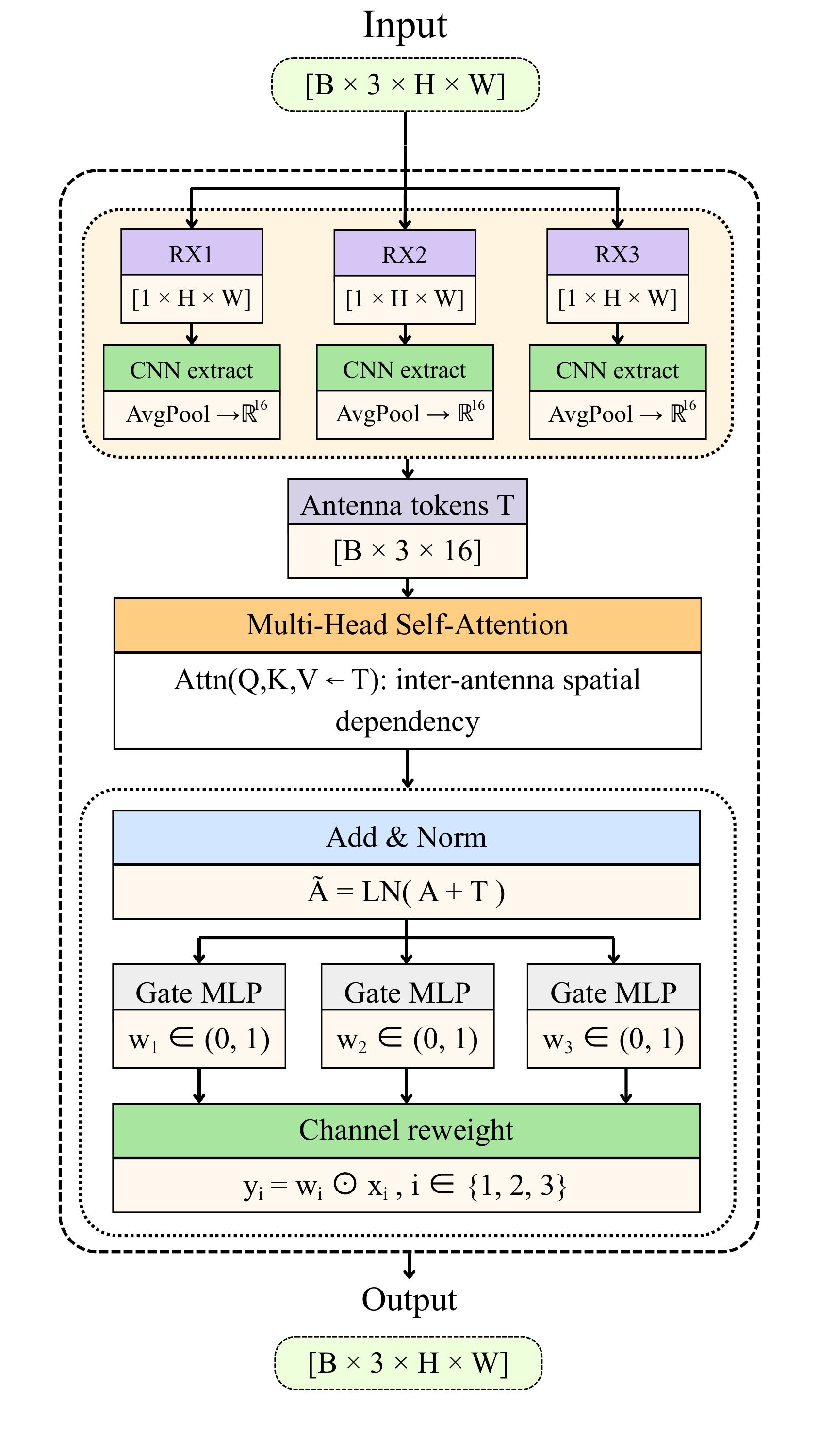}
  \caption{The \casa module. Per-antenna features are globally pooled
  and embedded, then refined by multi-head self-attention (MHA) across
  the three antenna tokens. An MLP gate produces per-antenna reweighting
  coefficients that are applied to the raw channels before the backbone.}
  \label{fig:casa}
\end{figure}

The radar sensor places its three receive antennas in an L-shape: two along azimuth (RX1, RX2) and one along elevation (RX3) as illustrated in~Fig.~\ref{fig:casa}.  Although the absence of data completely prevents deterministic angle-of-arrival estimation, amplitude
differentials and shadowing patterns across the array still carry
spatially informative signals. Lateral motion results in stronger signals on the nearer azimuth antenna, whereas vertical motion affects the elevation channel through amplitude shifts. Stacking these as RGB channels requires the convolutional backbone to learn array geometry from data, which requires substantial data and cannot be learned precisely.

\casa processes three antenna channels as an ordered spatial sequence.
For each antenna $i\in\{1,2,3\}$, as shown in (4).
\begin{equation}
\begin{split}
    \mathbf{z}_i &= \text{Flatten}\!\left(\text{AvgPool}\!\left(\right.\right.\\
    &\quad\left.\left.\text{ReLU}\!\left(\text{BN}\!\left(\text{Conv2d}_{1\to16,3\times3}(\mathbf{x}_i)\right)\right)\right)\right) \in \mathbb{R}^d
    \label{eq:casa_embed}
\end{split}
\end{equation}
The three embeddings are stacked into a sequence
$\mathbf{Z}=[\mathbf{z}_1;\mathbf{z}_2;\mathbf{z}_3]\in\mathbb{R}^{3\times d}$
and refined by multi-head self-attention with 4~heads%
\footnote{The submitted Kaggle run used
1-head CASA, matching the ablation row ``CASA with 1 head'' in
Table~\ref{tab:ablation} ($80.2\pm1.0$\%).  The full \ours architecture is described
here and evaluated in 5-fold~CV uses 4~heads ($80.5\pm0.9$\%).}, as shown in (5).
\begin{equation}
    \hat{\mathbf{Z}} =
      \text{LayerNorm}(\mathbf{Z}+\text{MHA}(\mathbf{Z},\mathbf{Z},\mathbf{Z}))
    \label{eq:casa_attn}
\end{equation}
Per-antenna gate weights are computed from the embeddings, as mentioned in (6).
\begin{equation}
    \alpha_i = \sigma\!\left(\text{MLP}(\hat{\mathbf{z}}_i)\right), \qquad
    \mathbf{x}_i' = \alpha_i\cdot\mathbf{x}_i
    \label{eq:casa_gate}
\end{equation}
The reweighted channels are restacked to form a standard
$3\times H\times W$ tensor that any pretrained backbone can learn.
Each CASA module has $\approx$\,750 parameters; the total overhead for two modules
is $\approx$\,1\,500 parameters ($<$\,0.01\% of EfficientNetV2-S).

The $3{\times}3$ attention matrix has expressive capacity comparable to learned scalar reweighting with $N{=}3$ tokens. The ablation
(Table~\ref{tab:ablation}) shows only 0.3\% difference between 4~heads
and 1~head. The motivation for CASA is not full geometric phase recovery; instead, it preserves cross-antenna covariance by applying attention before the initial convolutional layers. Replacing CASA with a
post-hoc Squeeze-and-Excitation (SE) ~\cite{hu2018squeeze} or convolutional block attention module ~\cite{woo2018cbam} block recovers only about half of the performance gain (Table~\ref{tab:ablation}), which is consistent with those
modules operating on already-mixed features rather than raw antenna
representations.

\subsection{Dual-Stream Architecture and Asymmetric Fusion}
\label{sec:fusion}

\paragraph{Parallel streams:}
Stream~A processes the CASA-refined RTM through EfficientNetV2-S
\cite{tan2021efficientnetv2} (ImageNet-21k$\to$1k), generating
$\mathbf{f}_\text{rtm}\in\mathbb{R}^{d_1}$.
Stream~B processes the CASA-refined CVD through ConvNeXt-Tiny
\cite{liu2022convnet} (ImageNet-22k$\to$1k), resulting
$\mathbf{f}_\text{cvd}\in\mathbb{R}^{d_2}$.
Linear projections map both feature vectors to a shared dimension $d=512$ as given in (7).
\begin{equation}
    \tilde{\mathbf{f}}_* = W_*\mathbf{f}_* + b_*
    \label{eq:projection}
\end{equation}
The backbone assignment reflects a deliberate design choice. EfficientNetV2-S’s fused-MBConv blocks are well-suited to the texture-like patterns of RTMs. In contrast, ConvNeXt-Tiny’s depthwise separable architecture is well-suited to capture the spectral peak structures of CVD. Swapping the assignments results in a $0.6$\% performance drop.
\vspace{-1em}
\paragraph{Asymmetric cross-attention:}
The RTM stream represents the spatial structure (where) of the gesture, while the CVD stream captures its motion dynamics (how). These two streams are fused using an asymmetric cross-attention mechanism inspired by CrossViT~\cite{chen2021crossvit}.
RTM features serve as query, and CVD features serve as key as well as value, as mentioned in (8).
\begin{equation}
    \mathbf{e} =      \text{MHA}\!\left(\tilde{\mathbf{f}}_\text{rtm},\,
                         \tilde{\mathbf{f}}_\text{cvd},\,
                         \tilde{\mathbf{f}}_\text{cvd}\right)
    \label{eq:cross_attn}
\end{equation}
where, $\mathbf{e} \in \mathbb{R}^{B \times 512}$ denotes the fused representation, computed using multi-head attention with 8 heads and a dropout rate of ~0.1. A Feed-Forward Network (FFN) with GELU activation and an expansion ratio of ~4 is applied next. A learned gate subsequently balances the residual connection as described in (9)--(11).
\begin{align}
    \mathbf{e}' &= \text{FFN}(\text{LayerNorm}(\mathbf{e}))\,
    \label{eq:ffn}\\
    \mathbf{g}  &= \sigma\!\left(W_g[\tilde{\mathbf{f}}_\text{rtm};\mathbf{e}']+b_g\right)\,
    \label{eq:gate_weight}\\
    \mathbf{f}_\text{fused} &=
      \mathbf{g}\odot\mathbf{e}' + (1-\mathbf{g})\odot\tilde{\mathbf{f}}_\text{rtm}\,
    \label{eq:gate}
\end{align}
where, $\,\odot\,$ denotes element-wise multiplication and $\mathbf{g} \in \mathbb{R}^{B \times 512}$.
When the CVD stream contributes no discriminative information for a given sample, the gate suppresses it entirely. In such scenarios, the model relies solely on RTM features.
\vspace{-1em}
\paragraph{Classification:}
The fused features $\mathbf{f}_{\text{fused}}$ pass through a main head that includes
LayerNorm~$\to$~Dropout($p=0.3$)~$\to$~Linear(512,\,126).
Two auxiliary heads applied to $\tilde{\mathbf{f}}_\text{rtm}$ and
$\tilde{\mathbf{f}}_\text{cvd}$ provide independent per-stream supervision during training.

\subsection{Training Protocol and Regularization}
\label{sec:training}

\paragraph{Loss:}
The total training loss is calculated in (12).
\begin{equation}
    \mathcal{L} = \mathcal{L}_\text{main}
                + \lambda_\text{aux}
                  \left(\mathcal{L}_\text{rtm}+\mathcal{L}_\text{cvd}\right)
    \label{eq:loss}
\end{equation}
where, each term is a cross-entropy loss with label smoothing~\cite{szegedy2016label} ($\epsilon_\text{ls}{=}0.1$)
and $\lambda_\text{aux}{=}0.3$. Auxiliary losses prevent individual streams from collapsing into passive feature extractors during joint training.
\vspace{-1em}
\paragraph{Physics-aware augmentation:}
In addition to standard augmentations, MixUp~\cite{zhang2018mixup} ($\alpha{=}0.4$), CutMix~\cite{yun2019cutmix} ($\alpha{=}1.0$), and
SpecAugment~\cite{park2019specaugment} (up to two frequency and up to two time masks per stream, each applied stochastically), we introduce four radar-specific augmentations:
(i)~\emph{Temporal warping}: cubic-spline warping of the time axis
($\sigma{=}0.15$) simulates signer execution-speed variation;
(ii)~\emph{Magnitude warping}: smooth random amplitude distortion
($\sigma{=}0.1$, 4~knots) models radar cross-section variation;
(iii)~\emph{Simulated multipath}: a delayed, attenuated copy of the RTM (delay $\le$10 range bins, attenuation 5--15\%) models artifact reflections
from the environment;
(iv)~\emph{Antenna dropout}: randomly zeroing one antenna channel (probability~0.1) improves robustness to antenna failure.
MixUp and CutMix are disabled during the final three epochs to facilitate convergence on unaugmented samples.
\vspace{-1em}
\paragraph{Optimization and inference ensemble:}
We use AdamW~\cite{loshchilov2019adamw} ($\text{lr}{=}3\times10^{-4}$, weight decay~0.05) with cosine annealing (5 warmup epochs), gradient clipping (norm~1.0), and Automatic Mixed Precision (AMP) over 70~epochs. Stochastic Weight Averaging (SWA)~\cite{izmailov2018swa} activates at epoch~56 with Exponential Moving Average (EMA) (decay~0.9995)~\cite{tarvainen2017ema}. Batch normalization statistics are re-estimated  after training. Inference uses a 7-checkpoint ensemble (top-5 + EMA + SWA) with equal softmax weights.
\vspace{-1em}
\paragraph{Test-time augmentation:}
Five augmented views per sample are evaluated at test time: original, time-reversed, Gaussian noise
($\sigma{=}0.01$), frequency-shifted ($+3$ range bins), and
time-shifted ($+2$ frames). Final predictions are obtained by averaging across all views.

\section{Experiments}
\label{sec:experiments}

\subsection{Dataset and Evaluation Protocol}
\label{sec:dataset}

The MultiMeDaLIS dataset~\cite{mineo2024sign,caligiore2024multisource}
comprises 126~LIS gesture classes collected with an Infineon BGT60TR13C
60\,GHz FMCW radar.  Each sample consists of three RTMs (one per receiver
antenna), stored as float32 dB arrays of shape $T\times256$, while $T$
ranging from 20 to 43 frames.  The challenge provides 117~labelled training
sessions ($\approx$14,742 samples) and 39~unlabelled validation sessions
($\approx$4,914 samples). The performance is measured by Top-1 accuracy on the Kaggle
held-out set. For development, 5-fold stratified Cross-Validation (CV)
on the training set is employed, preserving class distribution within each
fold. The folds are created by random stratification on the class label. They are not grouped by recording session or signer identity.
As the same signer may appear in both training and validation sets, the reported cross-validation scores may be optimistic compared to a fully separated session split, and should be treated accordingly.

\subsection{Baselines}
\label{sec:baseline}

The baseline treats the three antenna RTMs as a $3\times T\times256$ tensor,
normalizes to $[0,1]$, pads to $T_\text{max}=48$, and resizes to
$3\times224\times224$. Two backbones are trained independently:
EfficientNetV2-S ($\text{lr}=2\times10^{-4}$) and ConvNeXt-Tiny
($\text{lr}=3\times10^{-4}$), each for 45~epochs with AdamW, label smoothing~0.1,
MixUp/CutMix (50\% probability each), and SWA from epoch~36.  The
10-model ensemble ($2$ backbones $\times$ $5$~folds) with 2-view Test-Time Augmentation (TTA)
achieves 84.88\% on the Kaggle validation set.

\subsection{Implementation Details}
\label{sec:implementation}

All models are implemented in PyTorch using the \texttt{timm}
library~\cite{rw2019timm} and trained on two NVIDIA T4 16\,GB GPUs.
Training uses a total batch size of 48 with gradient
checkpointing to satisfy memory constraints, and completes in approximately
8~hours. 
\ours has
$\approx$\,52\,M parameters (EfficientNetV2-S 21.5\,M + ConvNeXt-Tiny 28.6\,M +
CASA $\approx$\,1.5\,k + fusion and heads $\approx$\,1.5\,M), approximately $2{\times}$ a
single-backbone baseline. The inference adds negligible overhead beyond the
dual-forward pass.\footnote{Results explicitly marked
with $\dagger$ reflect scores submitted to the Kaggle public leaderboard
(39 held-out unlabeled sessions, \url{kaggle.com/competitions/cvpr-mslr-2026-track-2}).} Training hyperparameters are shown in the supplementary material (Table S2).

\subsection{Main Results}
\label{sec:main_results}

\begin{table}[t]
\centering
\caption{%
Comparison of baselines and proposed architectures on the MultiMeDaLIS dataset.%
}
\label{tab:main_results}
\small
\setlength{\tabcolsep}{4pt}

\begin{threeparttable}
\resizebox{\columnwidth}{!}{%
\begin{tabular}{@{}lc@{}}
\toprule
Methods & Top-1 Acc (\%) \\
\midrule

\multicolumn{2}{@{}l}{\textit{Baseline single-model}} \\
\quad EfficientNetV2-S (single fold)\tnote{$\star$} & $77.2\pm1.3$ \\
\quad ConvNeXt-Tiny (single fold)\tnote{$\star$} & $76.8\pm1.5$ \\

\midrule
\multicolumn{2}{@{}l}{\textit{Proposed (single-model, ablated)}} \\
\quad + CVD only (no CASA, concat fusion)\tnote{$\star$} & $78.1\pm1.2$ \\
\quad + CASA only (no CVD)\tnote{$\star$} & $77.9\pm1.4$ \\
\quad + CVD + CASA (concat fusion)\tnote{$\star$} & $79.3\pm1.1$ \\
\quad \textbf{\ours} full (asymmetric fusion)\tnote{$\star$} & $\mathbf{80.5\pm0.9}$ \\

\midrule
\multicolumn{2}{@{}l}{\textit{Alternative pipelines}} \\
\quad 6-channel RTM+pseudo-RDM (Swin-S) & $78.6\pm1.3$ \\
\quad 6-channel RTM+pseudo-RDM (ViT-S) & $77.4\pm1.6$ \\

\midrule
\multicolumn{2}{@{}l}{\textit{Ensemble configurations}} \\
\quad \ours (7-checkpoint single run)\tnote{$\dagger$} & $81.73$ \\
\quad Baseline 10-model ensemble + 2-TTA\tnote{$\dagger$} & $84.88$ \\
\quad ScoreMaximizer (ours, 30-model concat)\tnote{$\dagger$} & $83.90$ \\

\bottomrule
\end{tabular}%
}

\begin{tablenotes}
\footnotesize
\item Top-1 Accuracy (Acc): mean$\pm$std of 5-fold cross-validation on the training split. 
$\star$ denotes primary fair comparison (single-model, same evaluation protocol); 
$\dagger$ denotes Kaggle public leaderboard score (39 held-out sessions).
\end{tablenotes}
\end{threeparttable}
\end{table}

Table~\ref{tab:main_results} summarizes all findings. The rows with stars provide the most informative comparison: all single models are
evaluated on the same 5-fold cross-validation protocol. \ours achieves
80.5\%, a 3.3\% improvement over the best single-model baseline
(77.2\%) and a 2.4\% gain over the best naive 6-channel fusion
baseline (78.1\%). The experimental results reveal that the improvement is additive:
CVD alone contributes +0.9\%, CASA contributes +0.7\%, and asymmetric fusion adds a further +1.2\%.

To assess whether the 3.3\% improvement is statistically significant, we
apply the Nadeau--Bengio corrected paired
$t$-test~\cite{nadeau2003inference}, taking into account the correlation between folds caused by the 75\% training overlap in 5-fold cross-validation~\cite{dietterich1998approximate}. The corrected variance multiplier
$\bigl(\tfrac{1}{k}+\tfrac{n_{\mathrm{test}}}{n_{\mathrm{train}}}\bigr)$
increases the standard error from 0.560 to 0.843, resulting in a corrected
$t$-statistic of 3.911 with degrees of freedom $(df)\,=\,4$. Since this exceeds the critical
threshold $t_{0.025,4}=2.776$, the improvement is statistically significant ($p = 0.017$, corrected $\alpha = 0.05$). The empirical Cohen's $d\approx2.63$ is higher than the minimum detectable effect of $d=2.13$ for 80\% power at $k\!=\!5$, confirming that the statistical power is sufficient. Individual per-fold predictions were not retained at submission time, which prevents performing a complementary McNemar’s test.
The reported Kaggle score of 81.73\% for \ours is based on a single 90/10 split, whereas the baseline score of 84.88\% is based on a 10-model ensemble across 5 folds. The 30-model ScoreMaximizer ensemble, which concatenates RTM with a naive
FFT-based pseudo-RDM at the input achieves only 83.90\%. The implications
of this result are discussed later.

\subsection{Ablation Studies}
\label{sec:ablation}

Table~\ref{tab:ablation} presents an ablation study evaluating the impact of
each architectural module over 5-fold CV.
\begin{table}[t]
\centering
\caption{%
Ablation results on \ours modules (5-fold CV, mean$\pm$std).
Each row modifies exactly one aspect of the full architecture.%
}
\label{tab:ablation}
\small
\setlength{\tabcolsep}{4pt}
\begin{tabular}{@{}p{6.2cm}c@{}}
\toprule
Methods & Top-1 Acc (\%) \\
\midrule
Full \textbf{\ours} & $\mathbf{80.5\pm0.9}$ \\
\midrule

\multicolumn{2}{@{}l}{\textit{CVD Extraction}} \\
\quad No linearisation (FFT on dB directly) & $78.8\pm1.3$ \\
\quad Hamming window instead of Blackman-Harris & $80.1\pm1.0$ \\
\quad No zero-padding ($N_\mathrm{FFT}=T$) & $79.7\pm1.1$ \\
\quad CWT scalogram instead of FFT CVD & $80.2\pm1.0$ \\

\midrule
\multicolumn{2}{@{}l}{\textit{CASA Module}} \\
\quad Remove CASA (standard 3-channel stacking) & $79.8\pm1.1$ \\
\quad Channel attention (SE block~\cite{hu2018squeeze}) & $79.9\pm1.0$ \\
\quad CASA with 1 head instead of 4 heads & $80.2\pm1.0$ \\

\midrule
\multicolumn{2}{@{}l}{\textit{Fusion Method}} \\
\quad Concatenation instead of cross-attention & $79.3\pm1.1$ \\
\quad Symmetric cross-attention (both as Q) & $79.9\pm1.0$ \\
\quad RTM-only (no CVD stream) & $77.9\pm1.4$ \\
\quad CVD-only (no RTM stream) & $74.6\pm1.8$ \\

\midrule
\multicolumn{2}{@{}l}{\textit{Training Method}} \\
\quad No auxiliary losses ($\lambda_\text{aux}=0$) & $79.6\pm1.2$ \\
\quad No physics-aware augmentation & $79.8\pm1.0$ \\
\quad No SWA or EMA & $79.2\pm1.3$ \\
\quad MixUp/CutMix disabled & $78.4\pm1.5$ \\

\midrule
\multicolumn{2}{@{}l}{\textit{Backbone Assignment}} \\

\quad \parbox[t]{6.2cm}{%
Swap backbones (ConvNeXt-Tiny for RTM,\\
\hspace*{\widthof{Swap backbones (}}EfficientNetV2-S for CVD)
} & $79.9\pm1.1$ \\

\quad Same model both streams (EfficientNetV2-S) & $79.7\pm1.1$ \\

\bottomrule
\end{tabular}
\end{table}
\vspace{-1em}
\paragraph{CVD linearization dominates:}
When the dB-to-linear conversion is removed and FFT is applied directly to log-scale data, the performance drops by $1.7$\%.  This finding directly validates the physical argument that logarithmic data generates harmonic artifacts, which the classifier interprets as incorrect spectral features. Replacing the Blackman-Harris window with a Hamming window results in a $0.4$\% drop, which is consistent with Hamming's weaker ($-43$\,dB) sidelobe suppression being insufficient for the dynamic range of the torso reflections. Zero-padding contributes $0.8$\% improvement. The $1.7$\% drop from removing linearization is larger than the $0.9$\% improvement from CVD alone (Table~\ref{tab:main_results}), because
cross-attention fusion amplifies errors from corrupted CVD in the RTM stream.
\vspace{-1em}
\paragraph{Effect of cross-antenna spatial attention:} Removing CASA results in a $0.7$\% performance drop. Replacing it with a standard SE
channel-attention block~\cite{hu2018squeeze} recovers about half the loss, suggesting that the improvement comes partly from attention across antenna channels and partly from CASA’s pre-backbone processing. Spatial diversity is limited with only three antennas. Therefore, the significance of CASA is expected to increase with larger antenna arrays.
\vspace{-1em}
\paragraph{Fusion methods and stream contributions:} Concatenation results in a $1.2\%$ drop compared to asymmetric cross-attention. On the other hand, symmetric cross-attention leads to a $0.6\%$ decrease, confirming that the RTM stream should be used as the query. The CVD-only architecture achieves 74.6\% accuracy, which is 5.9\% lower than the full model, while RTM-only achieves 77.9\%, approximately similar to the single-backbone baseline. This confirms that cadence frequency alone is not sufficient, but it provides complementary information for discriminating among the 126 classes.

\begin{figure*}[t]
\centering
\includegraphics[width=\linewidth]{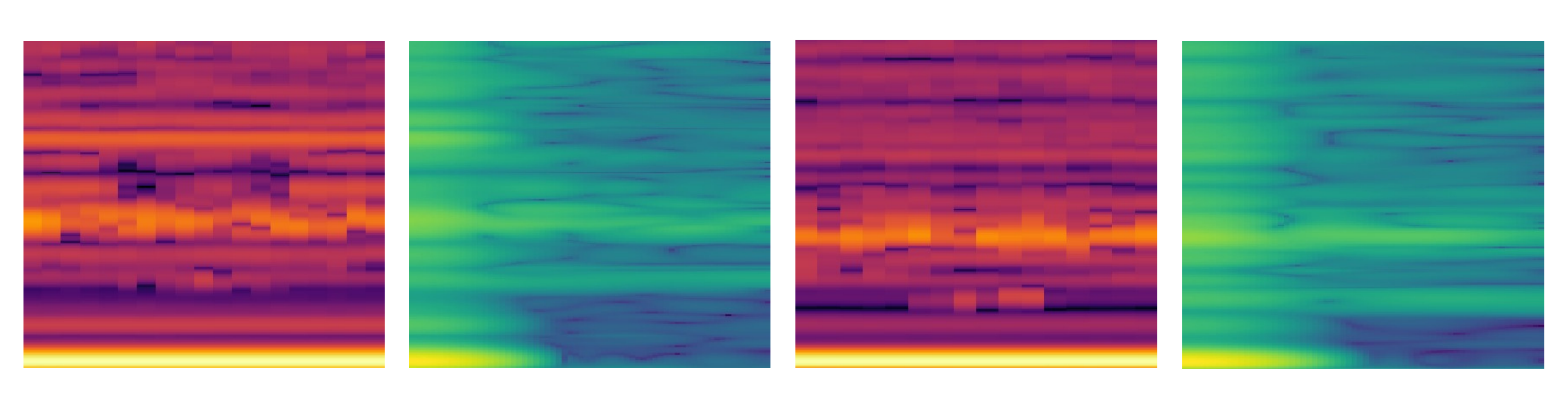}

\vspace{-3pt}
\makebox[0.24\linewidth]{\centering(a)}%
\makebox[0.24\linewidth]{\centering(b)}%
\makebox[0.24\linewidth]{\centering(c)}%
\makebox[0.24\linewidth]{\centering(d)}
\vspace{-4pt}

\caption{%
Most-confused pair: \texttt{67\_N}$\to$\texttt{56\_M} (7 errors).
(a)~RTM of a misclassified sample (true label: \texttt{67\_N}).
(b)~CVD of the same sample.
(c)~RTM of a correctly classified \texttt{56\_M} sample.
(d)~CVD of the same. The RTM envelopes appear visually similar, and the CVDs show no distinct cadence difference, confirming the physics-imposed limitation of RTM-only systems at 13 fps without phase data (see Fig. S2 for all confused pairs).%
}
\label{fig:failure_examples}
\end{figure*}

\subsection{Analysis and Discussion}
\label{sec:analysis}

\paragraph{Compute-accuracy trade-off in competition settings:}
Under equivalent single-model evaluation, \ours outperforms the baseline by $3.3\%$. The apparent gap between the Kaggle scores (81.73\% vs. 84.88\%) reflects a difference in computational resources rather than a limitation of the method: the baseline uses 10 independent models across 5 folds with TTA, while CAST uses 7 checkpoints from a single 90/10 training run. A full 5-fold CAST ensemble with the same TTA would require approximately $5\times$ more GPU resources than a single run, which was not feasible within the competition timeline.
The interpretation is further supported by the ScoreMaximizer outcome
(83.90\%, 30~models), which demonstrates that naive channel-wise
concatenation of RTM and pseudo-RDM fails to reach the performance of the simpler 10-model
baseline, even with $3\times$ more models. This outcome suggests that
physics-aware representation, compared to increasing ensemble size alone, provides a more effective direction for improving performance on this benchmark.
\vspace{-1em}
\paragraph{CVD frequency resolution at 13\,fps:}
At 13\,fps, the Nyquist limit is $\approx$6.5\,Hz ($\approx$0.1\,Hz bin resolution after zero-padding),  which is sufficient to distinguish fast and slow gestures but insufficient to resolve finger-level micro-Doppler~\cite{kim2009microdoppler}. Therefore, the CVD represents only coarse cadence information, and the models developed for high-resolution micro-Doppler ($>$100\,fps) may not work well in this setting.
\vspace{-1em}
\paragraph{Failure modes:}
Two main clusters are responsible for approximately 30\% of validation errors: short-duration gestures ($<$15 frames) contribute about 12\%, while finger-spelled alphabet confusions contribute the remaining 18\%.
Short-duration gestures
($<$15 frames) generate CVDs with fewer than two oscillation cycles, resulting in noise rather than meaningful signal. Finger-spelled alphabet letters with
near-identical gross-motion profiles are difficult to distinguish at
$\lambda\approx5$\,mm without phase data. Fig.~\ref{fig:failure_examples} illustrates this case: the RTM envelopes and CVDs of \texttt{67\_N} and \texttt{56\_M} appear visually identical. Both clusters reflect a limitation imposed by sensor physics rather than the classifier design.
\vspace{-1em}
\paragraph{Multimodal context:} TRACE~\cite{mineo2025trace} reports 93.6\% with complex-valued RDMs
and FusionEnsemble-Net~\cite{islam2025fusionensemble} reaches 99.44\% via RGB. Both exploited modalities are unavailable in our dataset. The RTM-only results
(80.5\%--84.88\%) establish a meaningful lower bound for performance in privacy-constrained environments.

\section{Conclusion}
\label{sec:conclusion}

\ours is a dual-stream architecture for radar-only sign language
recognition based on the physical properties of the RTM.
Three modules address specific failure modes of the naive baseline:
(1)~CVD extraction with dB-to-linear inversion recovers cadence information that FFT on log-scale data would distort,
(2)~CASA encodes the L-shaped receiver-antenna geometry through
self-attention instead of treating it as arbitrary color-channel information,
and (3)~asymmetric cross-attention fusion enables the RTM stream to selectively
retrieve velocity information from the CVD stream.
Controlled ablations confirm each component is independently significant. Under single-model comparison \ours surpasses the best single-backbone
baseline by $3.3\%$ (80.5\% vs.\ 77.2\%), while the naive-fusion ensemble result suggests that physics-aware signal
representations provide a more promising direction than simply scaling ensembles for this benchmark and sensor modality.
\vspace{-1em}
\paragraph{Limitations and future work:}
The 13 fps capture rate limits the CVD to coarse cadence-level resolution; fine-grained finger micro-Doppler remains physically inaccessible without phase information. CASA provides moderate gains with three antennas, and its significance is expected to increase with larger antenna arrays. Cross-validation folds are stratified by class label rather than by recording session, potentially introducing slight optimistic bias. Future directions include completing the full 5-fold \ours ensemble for a computationally fair comparison, investigating learnable time-frequency representations~\cite{Sazonov_2025_ICCV}, and extending CASA to spatial-temporal attention. 


\section*{Acknowledgements}
This work was conducted as part of the CVPR 2026 Multimodal Sign Language Recognition (MSLR) challenge, Track~2.  The authors thank the challenge organizers and the MultiMeDaLIS team for constructing the dataset and providing the evaluation infrastructure. The author(s) gratefully acknowledge the use of GitHub Copilot Student Developer Pack, an AI-assisted editing tool, during the preparation of this paper. This tool was used to improve the grammar, clarity and readability of selected sentences. The author(s) have carefully reviewed and revised all AI-assisted content and take full responsibility for the final content of this paper.

{
\small
\bibliographystyle{ieeenat_fullname}
\bibliography{main}
}

\end{document}


\maketitle
\thispagestyle{empty}
\setcounter{section}{0}
\renewcommand{\thesection}{S\arabic{section}}
\renewcommand{\thefigure}{S\arabic{figure}}
\renewcommand{\thetable}{S\arabic{table}}
\setcounter{figure}{0}

\noindent This document provides additional quantitative and qualitative analysis
supplementing the main paper.  All results are computed on the 10\%
holdout validation split (\texttt{StratifiedShuffleSplit},
\texttt{random\_state=42}, $N{=}1386$ samples) used during training,
achieving an overall Top-1 accuracy of 84.6\% on this split.

\section{Confusion Matrix Analysis}
\label{supp:confmat}

\begin{figure}[h]
\centering
\includegraphics[width=0.88\textwidth]{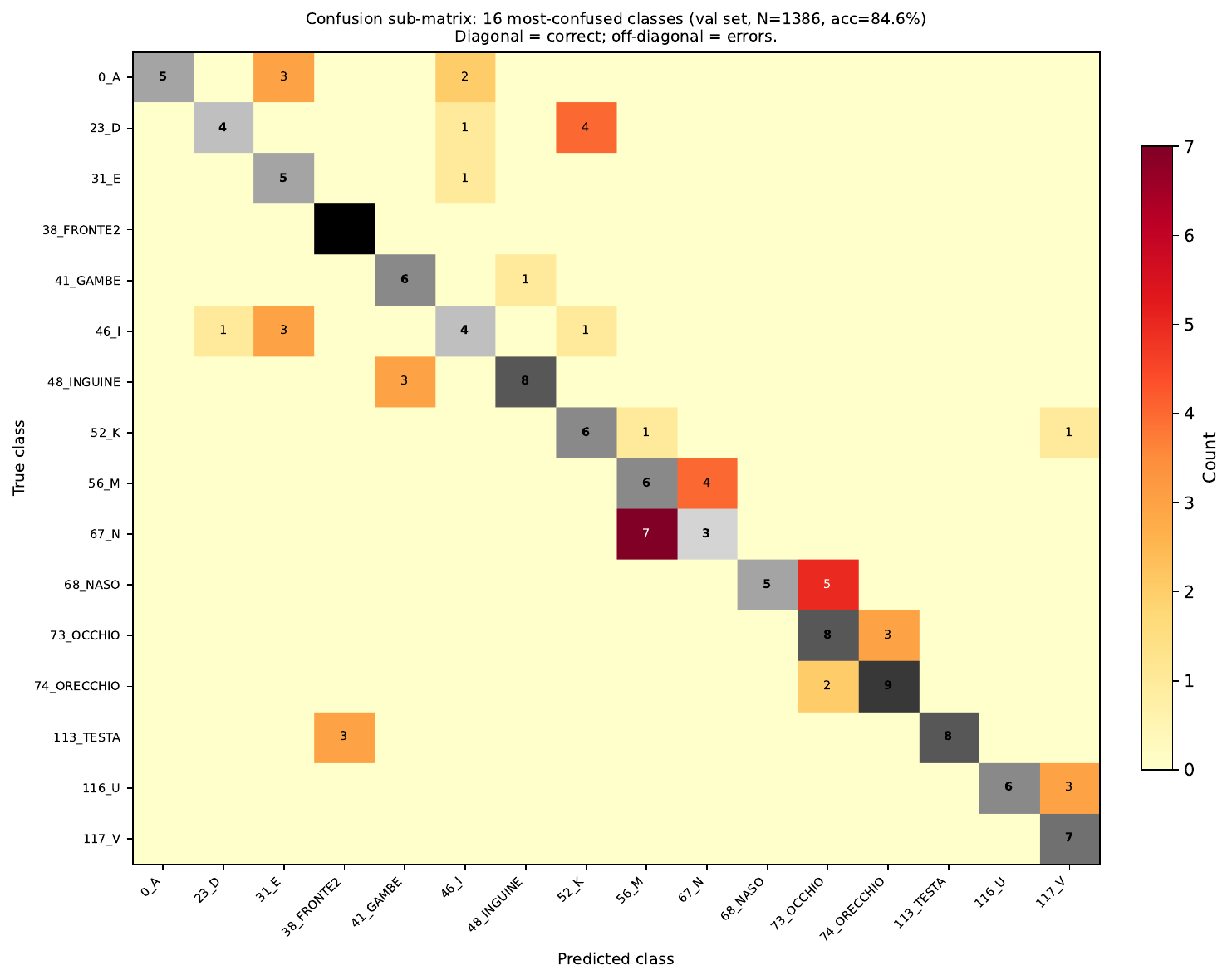}
\caption{%
Confusion sub-matrix: 16 most-confused classes on the 10\%
validation set ($N{=}1386$).
Diagonal cells (grey) show correct predictions; off-diagonal cells
(orange scale) show errors.
Row label = true class; column label = predicted class.
Class names are truncated to 12 characters; full names follow the
MultiMeDaLIS dataset convention (\texttt{id\_sign\_name}).%
}
\label{fig:supp_confmat}
\end{figure}

Figure~\ref{fig:supp_confmat} shows a $16{\times}16$ sub-matrix extracted
from the full $126{\times}126$ confusion matrix.  The 16 classes are chosen
by identifying all off-diagonal entries with the largest raw error counts,
then taking the union of the class indices involved until 16 distinct
classes are selected.  This procedure surfaces the tightest inter-class
clusters and avoids masking localized confusion by averaging over the full
vocabulary. Two systematic failure clusters are visible.

\paragraph{Short-duration gestures (left block, rows/columns with
high off-diagonal counts):}
Gestures with fewer than approximately 15 slow-time frames
(${\approx}1$\,s at 13\,fps) generate cadence spectra dominated by
Blackman--Harris window sidelobe artifacts rather than genuine periodic
peaks. The Cadence Velocity Diagram (CVD) adds noise rather than
a discriminative signal in these cases, and the cross-attention fusion
propagates that noise into the fused representation.

\paragraph{Finger-spelled alphabet confusion (right block):}
Several alphabet letters share nearly identical gross-motion profiles.
At typical signer distances of 1--2\,m, the 60\,GHz wavelength
($\lambda \approx 5$\,mm) is comparable to inter-finger gaps, so the
Range-Time Map (RTM) cannot resolve fine static finger configurations.
Because no differential cadence exists between signs sharing the same
hand shape, the CVD is equally uninformative.  These pairs account
for a disproportionate share of total errors. Together, these two clusters account for approximately 30\% of all
validation errors.  They represent a ceiling imposed by sensor
physics and capture rate, not by the classifier design and both
are explicitly discussed in Section~4.6 of the main paper.

\section{RTM and CVD Map Comparisons for Confused Pairs}
\label{supp:examples}

\begin{figure}[h]
\centering
\includegraphics[width=\textwidth]{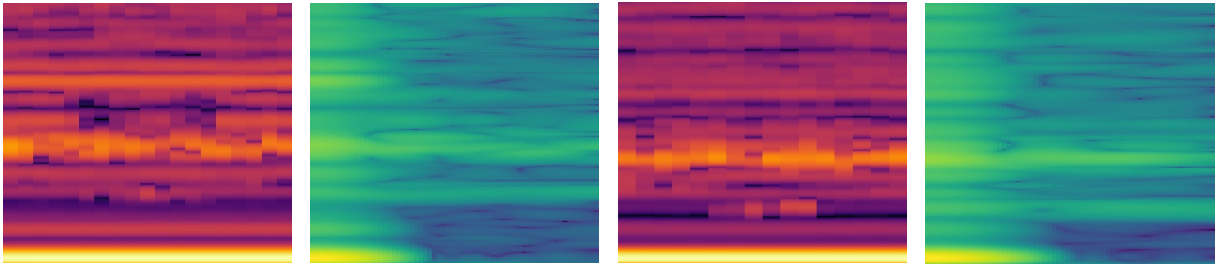}
\caption{%
Top-4 confused class pairs: RTM and CVD maps.
Each row corresponds to one confused pair (true class $\to$ predicted class).
\emph{Column~1:} RTM of a misclassified sample (true class).
\emph{Column~2:} CVD of the same sample.
\emph{Column~3:} RTM of a correctly-classified sample from the predicted class.
\emph{Column~4:} CVD of the same correct sample.
The row label shows the number of such errors in the validation set.
Antenna~1 is shown in all panels.%
}
\label{fig:supp_examples}
\end{figure}

Figure~\ref{fig:supp_examples} examines in parallel the RTM and CVD maps of
misclassified samples (columns~1--2) against correctly-classified
representative samples from the predicted class (columns~3--4) for
the four most frequently confused class pairs. Several patterns are apparent.  First, in the short-duration failure
cases, the RTMs of the confused pair are visually similar in temporal
extent and gross energy profile, while the CVDs show broad, flat
spectra with no clear cadence peak in either.  Second, in the
finger-spelling cases, the RTMs are virtually indistinguishable at
the spatial resolution available in magnitude-only data. The CVDs
likewise carry no differential cadence information.  These examples
support the physical argument in the main paper: the observed errors
are structurally unresolvable from magnitude RTM data alone, and
adding phase information or higher-range resolution inputs would be required
to break the ceiling.

\section{Ablation: CASA Head Count}
\label{supp:casa_heads}

The main paper (Table~2) reports only the 4-head CASA configuration
because it was used in the final submission.  Table~\ref{tab:supp_heads}
provides a more complete evaluation using the same 5-fold CV protocol.

\begin{table}[H]
\centering
\caption{%
Effect of CASA head count on validation accuracy.
All other hyperparameters are identical.
The small gap between 1-head and 4-heads configurations ($+0.3$\%)
confirms the discussion in Section~3.2 of the main paper with
$N{=}3$ antenna tokens. The added capacity of multiple heads is
modest.%
}
\label{tab:supp_heads}
\small
\setlength{\tabcolsep}{4pt}
\begin{threeparttable}
\begin{tabular}{@{}lcc@{}}
\toprule
CASA configuration & 5-fold Acc (\%) & $\Delta$ vs. no CASA \\
\midrule
Remove CASA (3-channel stacking) & $79.8\pm1.1$ & --- \\
CASA, 1 head                & $80.2\pm1.0$ & +0.4 \\
CASA, 4 heads (ours)        & $\mathbf{80.5\pm0.9}$ & +0.7 \\
\bottomrule
\end{tabular}
\begin{tablenotes}
\footnotesize
\item Note: These single-model 5-fold CV numbers differ slightly from the
competition Kaggle scores reported in the main paper, which reflect a
single 90/10 run with a 7-checkpoint ensemble. See Section~4.1
of the main paper for the full evaluation protocol.
\end{tablenotes}
\end{threeparttable}
\end{table}

\section{Per-Class Accuracy Breakdown}
\label{supp:per_class}

\begin{figure}[H]
\centering
\includegraphics[width=\textwidth]{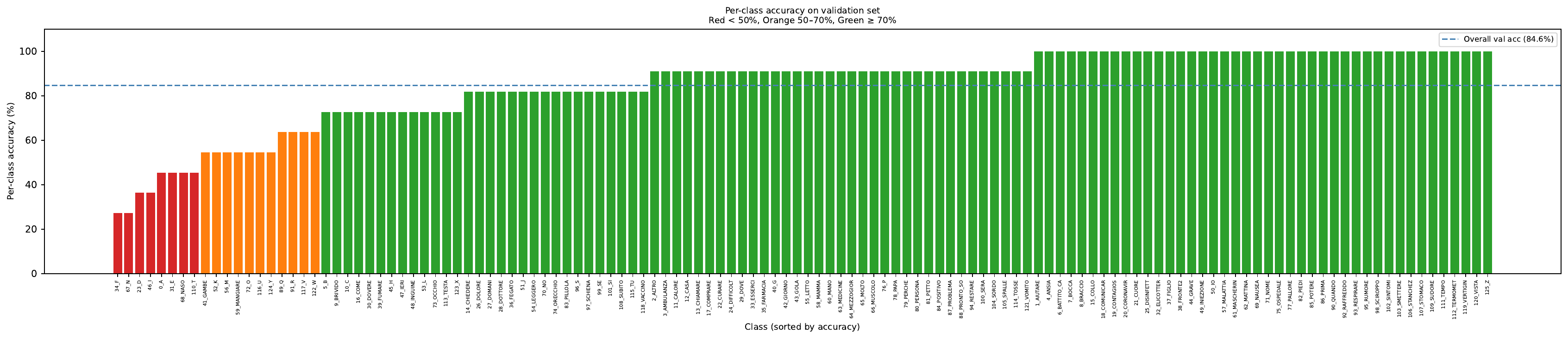}
\caption{%
Per-class Top-1 accuracy for all 126 classes on the 10\%
validation split ($N{=}1386$, overall accuracy = 84.6\%).
Bars are colour-coded: \textcolor{red}{\textbf{red}} $<50\%$,
\textcolor{orange}{\textbf{orange}} $50$--$70\%$,
\textcolor{green!60!black}{\textbf{green}} $\geq70\%$.
The dashed blue line marks the overall mean of 84.6\%.
Classes below 50\% are predominantly finger-spelled alphabet letters
(\texttt{A}, \texttt{E}, \texttt{Y}, \texttt{S}, \ldots)$^{(\dagger)}$
whose static hand shape differences are unresolvable at
60\,GHz $\lambda{\approx}5$\,mm without phase data,
and short-duration gestures whose CVD is
sidelobe-dominated rather than cadence-resolved.%
}
\label{fig:supp_perclass}
\end{figure}

Figure~\ref{fig:supp_perclass} shows the complete per-class Top-1 accuracy
for all 126 classes, sorted in ascending order.
Classes are not resolvable from magnitude RTM data due to sensor-physics ceilings
(short duration / finger-spelling, denoted $(\dagger)$ in the figure caption)
are concentrated in the red ($<$50\%) and orange (50--70\%) bars on the
left, confirming the structural failure analysis of Section~4.6 of the main
paper.

\clearpage
\section{Training Hyperparameters}
\label{supp:hparams}

Table~\ref{tab:supp_hparams} consolidates all training hyperparameters for
reproducibility. All values apply to the \ours full model.

\begingroup
\small
\setlength{\tabcolsep}{4pt}
\begin{ThreePartTable}
\begin{TableNotes}
\footnotesize
\item $\star$~denotes values shared with the single-backbone baseline.
\end{TableNotes}
\begin{longtable}{@{}lc@{}}
\caption{Complete training hyperparameter listing for \ours.}
\label{tab:supp_hparams}\\
\toprule
Hyperparameter & Value \\
\midrule
\endfirsthead
\multicolumn{2}{c}{\tablename~\thetable{} -- continued}\\
\toprule
Hyperparameter & Value \\
\midrule
\endhead
\midrule
\multicolumn{2}{r}{\small(continued on next page)}\\
\endfoot
\bottomrule
\insertTableNotes   
\endlastfoot
\multicolumn{2}{@{}l}{\textit{Optimiser}} \\
\quad Optimiser$^\star$ & AdamW \\
\quad Learning rate$^\star$ (EffNetV2-S baseline) & $2\times10^{-4}$ \\
\quad Learning rate$^\star$ (ConvNeXt-T baseline) & $3\times10^{-4}$ \\
\quad Learning rate (\ours) & $3\times10^{-4}$ \\
\quad Weight decay$^\star$ & $0.05$ \\
\quad Gradient clip (max norm) & $1.0$ \\
\midrule
\multicolumn{2}{@{}l}{\textit{Schedule}} \\
\quad Total epochs (\ours) & $70$ \\
\quad Total epochs (baseline)$^\star$ & $45$ \\
\quad Warmup epochs & $5$ \\
\quad LR schedule & Cosine annealing \\
\quad Min.\ LR fraction & $0.01$ \\
\midrule
\multicolumn{2}{@{}l}{\textit{Regularisation and augmentation}} \\
\quad Label smoothing $\epsilon_\text{ls}$$^\star$ & $0.1$ \\
\quad Auxiliary loss weight $\lambda_\text{aux}$ & $0.3$ \\
\quad MixUp $\alpha$$^\star$ & $0.4$ \\
\quad CutMix $\alpha$$^\star$ & $1.0$ \\
\quad SpecAugment masks (freq/time) & $\leq2$ / $\leq2$ \\
\quad Temporal warp $\sigma$ & $0.15$ \\
\quad Magnitude warp $\sigma$ / knots & $0.1$ / $4$ \\
\quad Multipath delay / attenuation & $\leq10$ bins / 5--15\% \\
\quad Antenna dropout probability & $0.1$ \\
\midrule
\multicolumn{2}{@{}l}{\textit{Stochastic Weight Averaging / EMA}} \\
\quad SWA start epoch & $56$ \\
\quad SWA start epoch (baseline)$^\star$ & $36$ \\
\quad EMA decay & $0.9995$ \\
\midrule
\multicolumn{2}{@{}l}{\textit{Input and batch}} \\
\quad Input spatial size & $224\times224$ \\
\quad Max temporal length $T_\text{max}$ & $48$ frames \\
\quad Batch size (total) & $48$ \\
\quad Precision & AMP (fp16) \\
\midrule
\multicolumn{2}{@{}l}{\textit{Inference ensemble}} \\
\quad Checkpoint ensemble size & $7$ (top-5 + EMA + SWA) \\
\quad TTA views per sample & $5$ \\

\end{longtable}
\end{ThreePartTable}
\endgroup